\documentclass{article}

\usepackage{spconf,amsmath,graphicx}
\usepackage{adjustbox,float,framed,multirow,subcaption,xcolor,booktabs,makecell,nicematrix,gensymb,mwe}
\usepackage[noadjust]{cite}
\usepackage[hidelinks]{hyperref}
\usepackage{enumitem}
\setlist{nosep, leftmargin=14pt}

\title{Utility of Pancreas Surface Lobularity as a CT Biomarker for Opportunistic Screening of Type 2 Diabetes}

\name{%
  \begin{tabular}{@{}c@{}} 
    Tejas Sudharshan Mathai\textsuperscript{$\star$},
    Anisa V. Prasad\textsuperscript{$\star$},
    Xinya Wang\textsuperscript{$\star$},
    Praveen T.S. Balamuralikrishna\textsuperscript{$\star$}, \\
    Yan Zhuang\textsuperscript{$\star$},
    Abhinav Suri\textsuperscript{$\star$},
    Jianfei Liu\textsuperscript{$\star$},
    Perry J. Pickhardt\textsuperscript{$\dagger$},
    Ronald M. Summers\textsuperscript{$\star$}
  \end{tabular}%
}

\address{%
  \textsuperscript{$\star$} Radiology and Imaging Sciences, National Institutes of Health Clinical Center \\
  \textsuperscript{$\dagger$} Department of Radiology, University of Wisconsin School of Medicine \&{} Public Health
}

\begin{document}

\maketitle

\begin{abstract}

\noindent
Type 2 Diabetes Mellitus (T2DM) is a chronic metabolic disease that affects millions of people worldwide. Early detection is crucial as it can alter pancreas function through morphological changes and increased deposition of ectopic fat, eventually leading to organ damage. While studies have shown an association between T2DM and pancreas volume and fat content, the role of increased pancreatic surface lobularity (PSL) in patients with T2DM has not been fully investigated. In this pilot work, we propose a fully automated approach to delineate the pancreas and other abdominal structures, derive CT imaging biomarkers, and opportunistically screen for T2DM. Four deep learning-based models were used to segment the pancreas in an internal dataset of 584 patients (297 males, 437 non-diabetic, age: 45$\pm$15 years). PSL was automatically detected and it was higher for diabetic patients (p=0.01) at 4.26 $\pm$ 8.32 compared to 3.19 $\pm$ 3.62 for non-diabetic patients. The PancAP model achieved the highest Dice score of 0.79 $\pm$ 0.17 and lowest ASSD error of 1.94 $\pm$ 2.63 mm (p$<$0.05). For predicting T2DM, a multivariate model trained with CT biomarkers attained 0.90 AUC, 66.7\% sensitivity, and 91.9\% specificity. Our results suggest that PSL is useful for T2DM screening and could potentially help predict the early onset of T2DM. 

\end{abstract}

\begin{keywords}
CT, Diabetes, Pancreas, Lobularity, Opportunistic Screening
\end{keywords}

\section{Introduction}
\label{sec:intro}

Type 2 Diabetes Mellitus (T2DM) is a chronic metabolic disease caused by a combination of insulin resistance and reduction in pancreatic islet beta cell function and volume \cite{Iwamoto2022,Macauley2015,Huang2025}. T2DM is projected to affect 700 million people worldwide by 2045 \cite{Sun2022}, with an estimated 8.7 million adults in the United States unaware that they have T2DM \cite{Fang2022}. T2DM can cause organ damage through an increased risk of vascular disease. Given the associated morbidity, early detection is critical for patient management. Currently, lab tests, such as an oral glucose tolerance test (OGTT), are used to diagnose T2DM. While incidental screening using CT is growing in popularity \cite{Tallam2022,Iwamoto2022,Remedios2025}, CT cannot be used for T2DM diagnosis due to high costs and radiation exposure. However, for those undiagnosed patients who may have undergone routine abdominal CT for other clinical indications, a positive screening result may inform the clinician to refer the patient for a confirmatory OGTT. 

\begin{figure*}[!htb]   

    \centering
    \includegraphics[width=\textwidth]{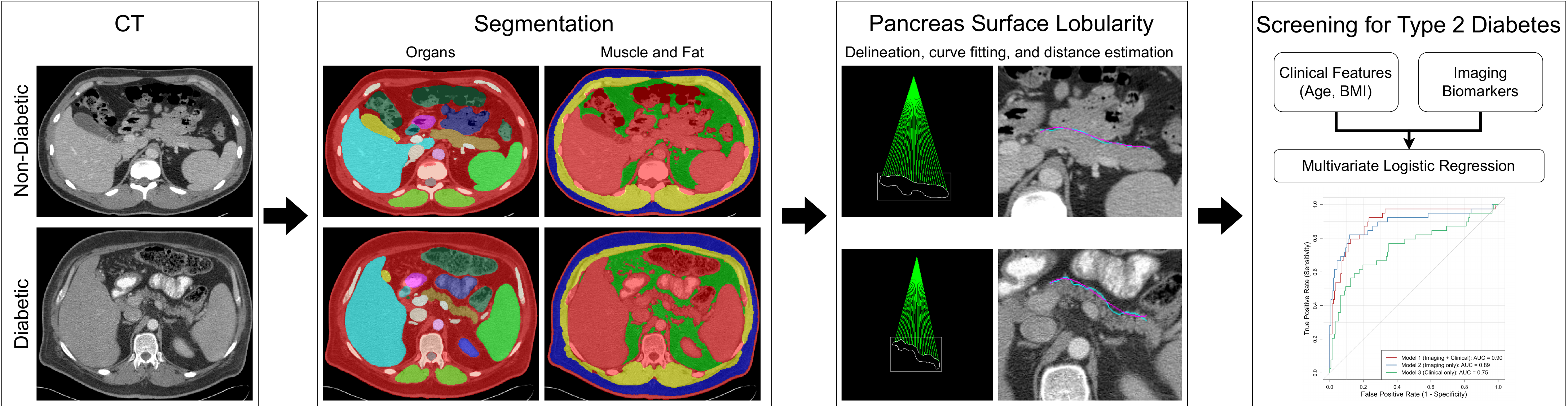}
    \caption{Fully automated framework for the diagnosis of T2DM. Top row shows a non-diabetic patient (male, 39 years) with a normal pancreas, while the bottom row shows a diabetic patient (male, 55 years) with an atrophied and lobular pancreas. A deep learning-based model segmented the pancreas (brown, left column) and other abdominal structures, while a previously validated approach segmented the muscle (yellow, right column), visceral fat (green) and subcutaneous fat (blue). The pancreas masks were then used to detect the anterior pancreas surface (cyan), a fourth-order polynomial (magenta) was fit to it, and the Euclidean distance between the two surfaces was measured to obtain the pancreas surface lobularity (PSL). Finally, PSL was combined with CT imaging and body composition biomarkers to diagnose T2DM. }
    \label{fig_money}
\end{figure*}

Prior studies have shown that the pancreas volume decreases by 7-22\% in patients with T2DM, along with increased fat deposition in the pancreas and its vicinity \cite{Iwamoto2022,Huang2025,Missima2025}. Morphological changes of the pancreas can be easily assessed on CT. Several studies have attempted to quantify such changes \cite{Tallam2022,suri_comparison_2024,Suri2025_diabetesPrediction,Remedios2025} with a few showing an association between insulin deficiency and pancreatic atrophy and loss of volume \cite{Williams2025,Lu2019}. In clinical practice with CT or MRI, the anterior pancreas surface in T2DM patients shows increased lobulations \cite{Gilbeau1992} or serrations \cite{Iwamoto2022} (see Fig. \ref{fig_money}). Very few studies have investigated this distinct feature corresponding to a lobulated pancreas surface \cite{Huang2025,Iwamoto2022,Sartoris2021,Macauley2015,Gilbeau1992}, and most relied on a manual or semi-automated measurement of pancreatic lobularity. Only two approaches \cite{Huang2025,Iwamoto2022} incorporated biomarkers of other organs and structures, such as the liver, visceral and subcutaneous fat, for prediction of T2DM.

In this pilot study, we propose a fully automated pipeline for the segmentation of the pancreas and other abdominal structures, followed by automatic estimation of pancreatic surface lobularity (PSL), and subsequent multivariate modeling for T2DM screening. Four deep learning-based segmentation models were evaluated for pancreas segmentation and PSL measurement. To our knowledge, we are the first to propose a fully automated approach to measure pancreatic lobularity and evaluate its utility for screening of T2DM.

\section{Methods}
\label{sec_methods}

\noindent
\textbf{Patient Sample.} In this retrospective study, 643 patients underwent abdominal contrast-enhanced CT exams at the National Institutes of Health (NIH) Clinical Center between 2005 – 2023. Abnormal glucose tests (e.g., HbA1c), test dates, medications taken, radiology reports, clinical notes, and admit/discharge diagnosis were extracted. 59 patients were excluded because the pancreas was not visible (as mentioned in the radiology report), underwent pancreatic surgery (either Whipple procedure or distal pancreatectomy), or HbA1c test results were unavailable. A total of 584 patients (297 males, 287 females, age: 45$\pm$15 years) were included as HbA1c lab tests and CT imaging was available. A patient was considered diabetic (n=147) if they had an HbA1c $\geq$ 6.5 and the CT exam was done after T2DM diagnosis, while they were non-diabetic (n=437) if they an HbA1c in the non-diabetic range before the CT exam and remained non-diabetic at least 4 years after the exam. Multiple CT scanners were used for imaging (GE, Siemens, Philips, Toshiba, and Canon) with a tube voltage between 90 - 140 kVp and patient-specific tube current settings. Only the portal-venous phase CTs were chosen. Volume dimensions ranged from 512 $\times$ 512 $\times$ (43-700) voxels and the spacing ranged between 1-5 mm. 

\medskip

\noindent
\textbf{Segmentation of Organs and Body Fat.} Figure \ref{fig_money} shows an overview. First, a previously validated tool \cite{Prasad2025} delineated 45 different structures in the body, and segmentations of the liver, pancreas, and spleen were extracted. This tool is henceforth called ``PancAP'', and it was chosen for its segmentation performance. PancAP is a 3D full-resolution nnU-Net trained on 1,350 CT volumes from the public PANORAMA dataset \cite{Alves2024}, and incorporated anatomy priors (e.g., skeleton, lungs, kidneys) derived from TotalSegmentator (TS) \cite{wasserthal2023totalsegmentator}. The main purpose for PancAP was to segment the pancreas and its vasculature. Additionally, segmentations of the muscle, visceral and subcutaneous fat were obtained using another previously validated body composition tool \cite{Liu2025}. It utilized mixed- and self-supervision along with transfer learning to train a 3D nnU-Net model for body composition analysis.  

\medskip

\noindent
\textbf{Automated Computation of Imaging Biomarkers.} Segmentations of the liver, pancreas, spleen, muscle and fat (both visceral and subcutaneous) were used to compute imaging biomarkers such as volume, CT attenuation (mean and standard deviation), and fat fraction. Fat fraction was computed by dividing the number of voxels corresponding to adipose tissue within a structure (e.g., liver) by the total number of voxels in that structure. Adipose tissue voxels have CT attenuation within the [-190, -30] HU range. Biomarkers for muscle were calculated at the L3 vertebral level, while those for visceral and subcutaneous fat were calculated at the L1 vertebral level. Body composition extracted at these levels are widely used indicators of obesity in patients \cite{Tallam2022}.   

\medskip

\noindent
\textbf{Pancreas Surface Lobularity (PSL).} Similar to a previously validated approach \cite{Mathai2025_LSN,Sartoris2021}, PSL was calculated on a slice-by-slice basis using the pancreas segmentation mask. First, the axial slice containing the highest number of voxels corresponding to the pancreatic body and tail was selected. This was based on prior observations indicating that these regions show the greatest degree of lobulation and surface serration \cite{Iwamoto2022,Huang2025,Sartoris2021,Macauley2015,Gilbeau1992}. Three slices adjacent to this slice in both the superior and inferior directions were also chosen (total of 7 slices) for computing PSL. As shown in Figure \ref{fig_money}, the centroid of the pancreas in each slice was determined and projected onto the first row of that slice. From this point, radial lines of length 512 pixels were cast across an angular range of 0$\degree$ to 180$\degree$ to intersect with the binary pancreas segmentation, thereby identifying the anterior pancreas surface. Points lying outside the segmentation were discarded, and those remaining were connected to form a continuous curve along the pancreas surface. The curve was also trimmed by 2mm on either side \cite{Mathai2025_LSN} to account for outliers. 

A fourth-order polynomial was fit to the detected surface to obtain a smooth approximation. For each pixel along the detected surface, the Euclidean distance (in millimeters) to the nearest point on the fitted curve was computed. The mean of these distances was subsequently scaled by a constant factor of 10 to amplify subtle variations in surface irregularity. The final PSL score for each CT scan was defined as the median of the slice-wise PSL scores, thereby minimizing the influence of outliers.

\medskip

\noindent
\textbf{Screening of Type 2 Diabetes Mellitus.} The dataset was divided into training ($\sim$70\%, n = 408 patients) and testing ($\sim$30\%, n = 176 patients) subsets. The training data subset had 108 diabetic and 300 non-diabetic patients, while the test data subset had 39 diabetic and 137 non-diabetic patients, respectively. Using the biomarkers from each model, several bootstrapped multivariate logistic regression models were built (``glm'' function, ``binomial'' family, stats package, RStudio v.2024.04.2). These multivariate models used various combinations of the clinical and imaging biomarkers as features to diagnose T2DM. The baseline multivariate model comprised of only the clinical features (age and BMI), while the imaging biomarkers model contained only imaging biomarkers (no clinical features). The effect of using both clinical features and imaging biomarkers was evaluated. 

\medskip

\noindent
\textbf{Comparisons.} In addition to PancAP, the publicly available TotalSegmentator (TS) \cite{wasserthal2023totalsegmentator}, PanSegNet \cite{Zhang2025_PanSegNet}, and a pancreas sub-region segmenter \cite{Zhuang2025_subRegion} were also used to segment the pancreas. PancAP, TS and PanSegNet segmented the whole pancreas, whereas the pancreas sub-region segmenter distinguished the head, body and tail of the pancreas separately. 

\medskip

\noindent
\textbf{Reference Standard.} To quantify the performance of all four segmentation models, the pancreas was manually annotated by a clinician (2+ years of experience) on 25 portal-venous phase CT scans. The annotations were further verified by a board-certified radiologist with 30+ years of experience. 

\medskip

\noindent
\textbf{Statistical Analysis.} Segmentation was evaluated with Dice similarity coefficient (DSC) and average symmetrical surface distance (ASSD) error. Performance differences between the models were assessed using a Friedman test, while a Wilcoxon signed-rank test (Bonferroni adjusted) was used for pairwise comparisons. Classification performance was assessed using AUC, sensitivity, and specificity. An AUC $<$ 0.7 was considered ineffective. Youden's index was set as the threshold to compute specificity and sensitivity. A bootstrap test (pROC package, RStudio v.2024.04.2) was used to compare the ROC curves from two models. A p-value $<.05$ was considered statistically significant.

\begin{table*}[!htb]
\centering
\caption{Results of screening for Type 2 Diabetes using multi-variate regression models. 95\% confidence intervals are also provided. Bold and underlined font indicates best and second-best results. Clinical features were age and body mass index (BMI), whereas imaging features were derived from the CT scan. } 
\smallskip
\begin{adjustbox}{max width=\textwidth}
\begin{tabular}{l|cc| r ccc}
\hline
        Model   & \multicolumn{2}{c}{Segmentation}      &  & \multicolumn{3}{c}{Classification} \\
                & Dice & ASSD (mm)                      & Features & AUC & Sensitivity & Specificity   \\
\hline
\multirow{3}{*}{TotalSegmentator} & \multirow{3}{*}{0.75 $\pm$ 0.16} & \multirow{3}{*}{2.19 $\pm$ 2.24} & Clinical Only    & 0.75 (0.64, 0.85)    & 35.8 (23.1, 51.3)	& \textbf{93.4} (89.1, 97.1) \\
&  &  & Imaging Only          & \underline{0.89} (0.81, 0.95) & \textbf{71.8} (56.4, 87.2) & \underline{92.7} (88.3, 96.4) \\
& & & Clinical + Imaging    & \textbf{0.90} (0.83, 0.95) & \underline{64.1} (48.7, 79.8) & 90.5 (85.4, 95.6) \\

\hline
\multirow{3}{*}{PanSegNet} & \multirow{3}{*}{\underline{0.78 $\pm$ 0.18}} & \multirow{3}{*}{2.26 $\pm$ 3.12} & Clinical Only          & 0.74	(0.64, 0.84) & 30.7 (17.9, 46.2) & \textbf{94.9} (91.2, 98.5) \\
&  &  & Imaging Only          & \underline{0.88} (0.81, 0.95) & \textbf{71.8} (56.4, 84.6) & \underline{91.9} (87.5, 96.3) \\
&  &  & Clinical + Imaging    & \textbf{0.90} (0.84, 0.95) & \underline{64.1} (48.7, 79.5) & 91.2 (86.0, 95.6) \\

\hline
\multirow{3}{*}{Sub-Region} & \multirow{3}{*}{0.77 $\pm$ 0.17} & \multirow{3}{*}{\underline{2.05 $\pm$ 2.39}} & Clinical Only         & 0.74 (0.63, 0.82) & 30.8 (17.9, 46.2) & \textbf{94.9} (91.2, 97.8) \\
&  &  & Imaging Only          & \underline{0.89} (0.81, 0.94) & \textbf{69.2} (53.8, 84.6) & \underline{91.9} (87.6, 95.6) \\
&  &  & Clinical + Imaging    & \textbf{0.90} (0.84, 0.95) & \underline{66.7} (51.3, 79.5) & 89.8 (84.7, 94.9) \\

\hline
\multirow{3}{*}{PancAP} & \multirow{3}{*}{\textbf{0.79 $\pm$ 0.17}} & \multirow{3}{*}{\textbf{1.94 $\pm$ 2.63}} & Clinical Only             & 0.75 (0.65, 0.84) & 35.9 (20.5, 51.3) & \textbf{93.4 } (88.3, 97.1) \\
&  &  & Imaging Only          & \underline{0.89} (0.81, 0.96) & \textbf{74.3} (58.9, 87.2) & 89.8 (84.7, 94.2) \\
&  &  & Clinical + Imaging    & \textbf{0.90} (0.83, 0.96) & \underline{66.7} (51.3, 79.6) & \underline{91.9} (87.6, 96.4) \\

\hline

\end{tabular}
\end{adjustbox}
\label{table_LR_results}
\end{table*}

\section{Results}
\label{sec_results}

\noindent
\textbf{Segmentation Results.} Table \ref{table_LR_results} describes the segmentation results. Amongst the four models, PancAP achieved the highest Dice score of 0.79 $\pm$ 0.17 and lowest ASSD error of 1.94 $\pm$ 2.63 mm. PanSegNet achieved the next best Dice score of 0.78 $\pm$ 0.18, but the ASSD error was higher (2.26 $\pm$ 3.12 mm) compared to SubRegion at 2.05 $\pm$ 2.39 mm. Statistically significant differences were seen between PancAP and SubRegion (p = 0.022), TS (p = 0.001), and PanSegNet (p = 0.03) for both Dice and ASSD error. Similarly, a significant difference was seen between SubRegion and TS for Dice score (p $<$ 0.001) but not for ASSD (p = 0.09).

\medskip

\begin{figure}[!tbp]
     \centering
     \begin{subfigure}[b]{0.23\textwidth}
         \centering
         \includegraphics[width=\textwidth]{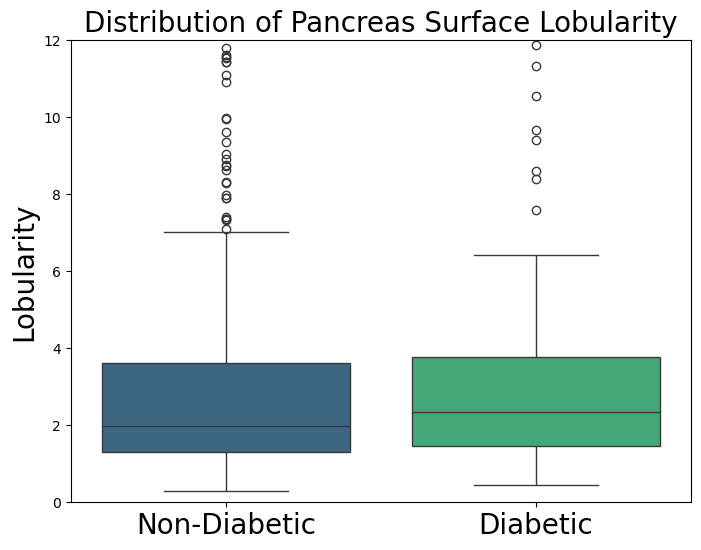}
     \end{subfigure}
     \begin{subfigure}[b]{0.23\textwidth}
         \centering
         \includegraphics[width=0.85\textwidth]{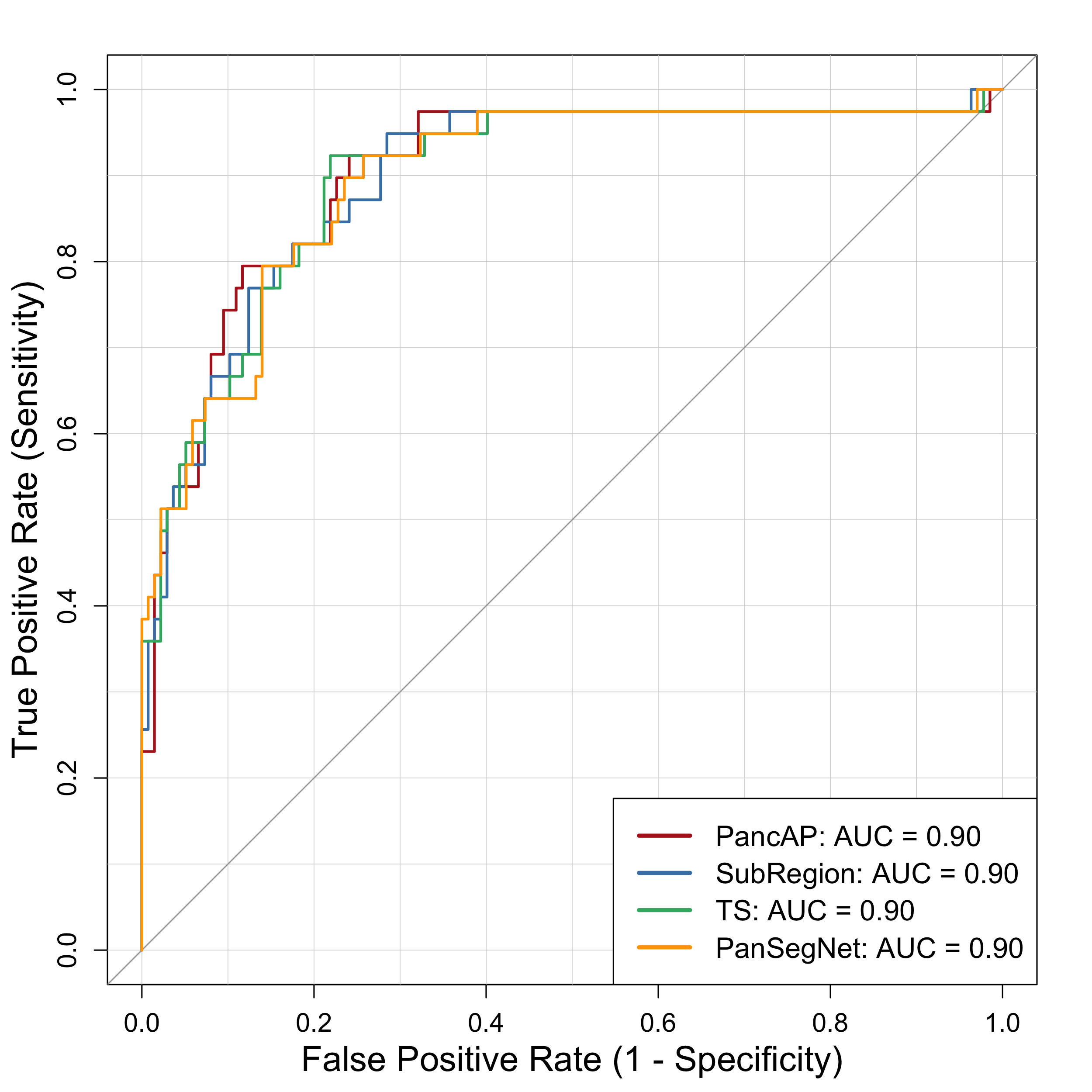}
         \label{fig:three sin x}
     \end{subfigure}
     
        \caption{Box plot (left) of pancreas surface lobularity scores for non-diabetic and diabetic patients by PancAP model. ROC curves for diagnosing Type 2 Diabetes with multivariate models incorporating clinical and imaging biomarkers.}
        \label{fig_plots}
\end{figure}

\noindent
\textbf{Classification Results.} As shown in Figure \ref{fig_plots}, PSL for diabetic patients was 4.26 $\pm$ 8.32, while PSL for non-diabetic patients was 3.19 $\pm$ 3.62 (p = 0.01). Comparing the multivariate regression models that used only clinical features (age and BMI) for screening of T2DM, the PancAP model achieved the highest AUC of 0.74 and sensitivity of 35.9\%. Both PanSegNet and the Sub-Region model attained specificity of 94.9\%, whereas PancAP obtained a slightly lower specificity of 93.4\%. Compared against the models that used only clinical features, the multivariate models that used imaging features alone showed significant improvement. PancAP again obtained the highest AUC of 0.89 and sensitivity of 74.3\%, albeit at a lower specificity of 89.8\%. 


Feature importance revealed that there were 16 imaging biomarkers that contributed to the diagnostic performance of the PancAP model: volume (liver, liver fat, pancreas, pancreas fat), mean CT attenuation (liver, spleen, and muscle at L3 level), standard deviation of CT attenuation (liver, spleen, pancreas, muscle at L3, visceral and subcutaneous fat at L1 level), fat fraction (pancreas, muscle at L3), and median PSL score. In this work, confounding variables were adjusted for such as volume of the liver and pancreas, liver fat content, body composition measures, age, and BMI. 

Combining both the clinical and imaging biomarkers together into a multivariate model lead to an AUC of 0.90 and specificity of 91.9\% by the PancAP model. However, the sensitivity decreased by $8$\% (66.7\% vs. 74.3\%). Across all segmentation models, compared against the clinical model, statistically significant differences were observed (p $<$ 0.01) when using multivariate models with either imaging biomarkers alone or a combination of clinical and imaging biomarkers. No statistically significant difference was seen between the combination model vs. the imaging biomarkers only model. Figure \ref{fig_plots} shows the receiver operating characteristics (ROC) AUC curves for the four multivariate regression models built from the four segmentation models. 






\section{Discussion}
\label{sec_discussion}

In this work, a fully automated pipeline was developed to delineate the pancreas, calculate the pancreatic surface lobularity, and combine it with other CT imaging biomarkers for the screening of T2DM. Previous literature has established that an acceptable level of overlap between the automated segmentation and the reference annotation was a Dice score greater than 75\% \cite{Prasad2025,Suri2025_diabetesPrediction,suri_comparison_2024}. All models that were evaluated in this work achieved this threshold. However, significant differences were seen (p $<$ 0.05) when comparing the PancAP model against the Sub-Region, PanSegNet, and TotalSegmentator models. Imaging biomarkers were then subsequently derived from the pancreas segmentations and the PSL was computed. As shown in Figure \ref{fig_plots}, the PSL for diabetic patients was higher compared to non-diabetic patients. This observation that diabetic patients have serrated pancreases with increased lobularity is corroborated by prior works \cite{Iwamoto2022,Huang2025}. 

The key difference in this work compared to prior approaches \cite{Iwamoto2022,Huang2025} is the fully automated measurement of pancreas surface lobularity. The best set of features to predict T2DM included both clinical features (age and BMI) and imaging biomarkers. In particular, volume, mean and standard deviation of CT attenuation for the liver, spleen, pancreas, muscle and body fat were identified. There was no significant difference between the combination model (imaging and clinical) and the imaging biomarkers only model. The results for screening of T2DM presented in this work (AUC of 90\%) are similar to the results in prior work \cite{Suri2025_diabetesPrediction,Huang2025}. 

Our results have many clinical implications. An atrophied and lobular pancreas with deep incisures is a known indicator of metabolic changes in the pancreas \cite{Huang2025,Iwamoto2022,Sartoris2021,Macauley2015,Gilbeau1992}. Prior work hypothesized that  inflammation, high blood glucose, and fatty infiltration of the pancreas leads to reduction in pancreas volume, increased fibrosis between lobules, and concomitant increase in surface lobularity \cite{Huang2025}. PSL is more evident during earlier stages of pancreatic disease \cite{Huang2025,Sartoris2021} and may act as a biomarker for early disease onset. Since there are $\sim$8 million adults in the United States with undiagnosed diabetes, there is a benefit to assessing pancreas lobularity as these patients can be potentially screened early and scheduled for a confirmatory diagnosis of pre-diabetes and T2DM. 

There are several limitations to this work. First, adjustment for sex was not done in this work. Men have increased lobulation and they are more prone to T2DM compared to women \cite{Huang2025,Sartoris2021}. Second, the success of the PSL measurement depends heavily on the pancreas segmentation performance of a model. Incorrect delineations of the pancreas can significantly affect that measurement process. While PancAP, SubRegion and TS succeeded in segmenting the pancreas for all scans, PanSegNet failed to segment the pancreas for 1 patient scan and it had to be excluded from the analysis. Third, PSL was only measured for the entire pancreas, and individual sub-regions (head, body, and tail) were not evaluated in this pilot work. Finally, this was a single-center retrospective study and further research needs to be done with large patient cohorts to validate the effectiveness of PSL. 

In conclusion, a fully automated approach was proposed to compute pancreas surface lobularity and combine it with clinical and CT imaging biomarkers for prediction of T2DM. The proposed approach can be easily extended to non-contrast CT and MRI. Pancreatic lobularity may be a useful indicator of early onset of disease, such as pre-diabetes, Type 2 Diabetes, and pancreatitis.

\clearpage

\section{ACKNOWLEDGEMENTS}      

\noindent
This work was supported by the Intramural Research Program of the NIH Clinical Center (project number 1Z01 CL040004). This work utilized the computational resources of the NIH HPC Biowulf cluster.

\section{Compliance with Ethical Standards}

This study was approved by the Institutional Review Board (IRB) at the NIH and performed with retrospectively acquired patient data. The need for informed consent was waived.

\bibliographystyle{IEEEbib}

\bibliography{isbi26_references}

\end{document}